\documentclass[letterpaper, 10 pt, conference]{ieeeconf}  

\IEEEoverridecommandlockouts                              

\usepackage[tracking=true]{microtype}
\usepackage{amsmath} 
\usepackage{amssymb}  
\usepackage{graphicx}
\usepackage{epsfig} 
\usepackage{mathptmx} 
\usepackage{times} 
\usepackage{amsmath} 
\usepackage{amssymb}  
\usepackage{xspace}
\usepackage{url}
\usepackage{float}

\usepackage[dvipsnames]{xcolor}
\usepackage[font={footnotesize}]{caption}
\usepackage{booktabs} 
\usepackage{tikz}
\usetikzlibrary{positioning,shapes.geometric,arrows,fit,shapes.symbols,svg.path,tikzmark,calc}
\usepackage{textcomp}
\usepackage{listings}
\usepackage{subcaption}
\pgfdeclarelayer{background}
\pgfdeclarelayer{foreground}
\pgfsetlayers{background,main,foreground}
\usepackage{siunitx}
\usepackage{adjustbox}
\usepackage{array}
\usepackage{multirow}

\usepackage{siunitx}

\let\labelindent\relax 

\usepackage[inline]{enumitem} 

\usepackage[backend=biber,
            url=false,
            isbn=false,
            doi=false,
            backref=false,
            style=ieee,
            natbib=true,
            mincitenames=1,
            maxcitenames=1,
            citestyle=numeric-verb,
            sorting=none,
            block=none]{biblatex}
\renewcommand{\bibfont}{\small}
\newcommand{\algname}{FogROS2-FT\xspace}
\newcommand{\sgc}{FogROS2-SGC\xspace}

\title{\LARGE \bf
FogROS2-FT: Fault Tolerant Cloud Robotics
}

\author{$^\dagger$Kaiyuan Chen$^{{1}}$,  Kush Hari$^{{1}}$, Trinity Chung$^{{1}}$,  Michael Wang$^{2}$, Nan Tian$^{2}$, Christian Juette$^{2}$, \\Jeffrey Ichnowski$^{3}$,  Liu Ren$^{2}$, John Kubiatowicz$^{1}$, Ion Stoica$^{1}$, and Ken Goldberg$^{1, 4}$ 
\thanks{$^{1}$Department of Electrical Engineering and Computer Science}%
\thanks{$^{2}$Robert Bosch Research and Technology Center North America, Sunnyvale, CA, USA}%
\thanks{$^{3}$Robotics Institute, Carnegie Mellon University}%
\thanks{$^{4}$Department of Industrial Engineering and Operations Research}%
\thanks{$^{1,4}$University of California, Berkeley, CA, USA }%
\thanks{$^\dagger$For correspondence and questions: {kych@berkeley.edu}}
}

\begin{document}

\maketitle
\thispagestyle{empty}
\pagestyle{empty}

\begin{abstract}
    Cloud robotics enables robots to offload complex computational tasks to cloud servers for performance and ease of management. 
However, 
cloud compute can be costly, 
cloud services can suffer occasional downtime, and connectivity between the robot and cloud 
can be prone to 
variations in network Quality-of-Service (QoS). 
We present FogROS2-FT (Fault Tolerant) to mitigate these issues by introducing
%
a multi-cloud extension that automatically replicates independent stateless robotic services
, routes requests to these replicas, and directs the first response back. With replication, robots can still benefit from cloud computations even when a cloud service provider is down or there is low QoS.  Additionally, many cloud computing providers offer low-cost ``spot'' computing instances that may shutdown unpredictably. Normally, these low-cost instances would be inappropriate for cloud robotics, but the fault tolerance nature of FogROS2-FT allows them to be used reliably. 
We demonstrate FogROS2-FT fault tolerance capabilities in 3 cloud-robotics scenarios in simulation (visual object detection, semantic segmentation, motion planning)
and 1 physical robot experiment (scan-pick-and-place).
Running on the same hardware specification,  
\algname achieves motion planning with up to 2.2x cost reduction and up to a 5.53x reduction on 99 Percentile (P99) long-tail latency.
\algname reduces the P99 long-tail latency of object detection and semantic segmentation by  2.0x and 2.1x, respectively, under network slowdown and resource contention. 
Videos and code are available at 
\url{https://sites.google.com/view/fogros2-ft}.
\end{abstract}

\section{Introduction}

\begin{figure}
    \centering
    \includegraphics[width=\linewidth]{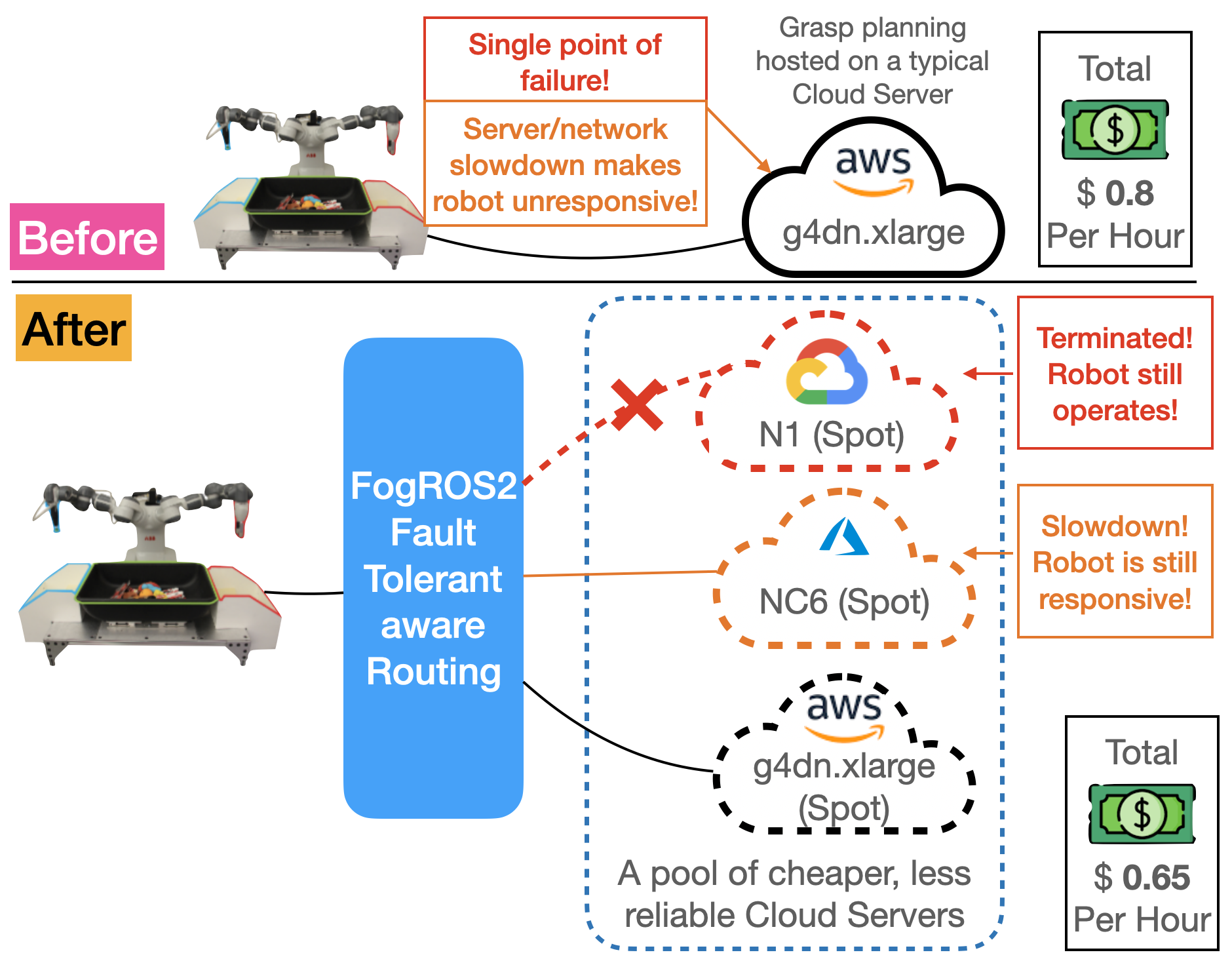}
    \caption{\textbf{\algname Overview.} \textbf{(Top)} Cloud robotics applications, such as grasp planning, when deployed on a single cloud server become a single point of failure.
    \textbf{(Bottom)} Instead, \algname provides a cost-efficient and fault-tolerant solution that deploys unmodified ROS2 applications to multiple low-cost cloud servers, making cloud-robotics applications resilient to individual server termination and network slowdowns.
    }
    \label{fig:intro:use_case}
\end{figure}

The complexity of foundational models~\cite{mildenhall2021nerf,kirillov2023segment,Rashid24lifelonglerf} and sophisticated robot algorithms~\cite{ichnowski2020fog,tanwani2019fog} exceed most onboard robot computing capabilities.
Cloud robotics provides shared access to on-demand resources and services with boosted performance and simplified management, enabling the deployment of compute-intensive algorithms on low-cost, mobile robots without powerful on-board hardware, such as GPU, TPU, and high-performance CPU.
In previous research, we developed FogROS2, which enables unmodified robotics code in Robot Operating System 2 (ROS2) to offload heavy computing modules to
an independent set of 
cloud hardware resources and accelerators.
FogROS2 used \emph{on-demand} servers that guarantee dedicated computing resources with high uptime (e.g., 99.99\,\%~\cite{aws_sla}). However, the network quality of service (QoS) between robots and the cloud can vary, and during rare cloud outages, robots lose all cloud-computing benefits.
Additionally, as on-demand instances can be expensive, many cloud providers offer \emph{spot VMs}\footnote{In Google Cloud Platform (GCP) and Microsoft Azure, these are called \emph{Spot Virtual Machines}. In Amazon Web Services, these are called \emph{Spot Instances}. More generally, they are also known as \emph{preemptible} or \emph{transient} machines.}  
at a significantly reduced price with the caveat that they can shut down unpredictably---making them (without fault tolerance) unsuitable for many robotics applications.
In this work, we introduce \algname, a fault-tolerant extension to FogROS2~\cite{ichnowski2023fogros} 
that provides robust performance against variable network QoS, infrastructure unavailability, and stochasticity of the robotic algorithms,
increasing the reliability and responsiveness of cloud robotics. 
By adding redundancy to cloud computation, we enable cloud-robotics tasks to continue operating effectively despite the following failures: 
 


\begin{enumerate}[label=(\Alph*),itemindent=3em,leftmargin=0pt]
\item \textit{Resource Unavailability}: Although cloud services have high uptime and are managed by dedicated experts, outages can still occur.
For example, an AWS outage affected the availability of the iRobot  applications~\cite{irobotaws}.  In addition, spot VMs may shut down unpredictably. 

\item \textit{Resource Oversubscription}: The cloud enables flexible usage of computational resources.
For example, one can oversubscribe to a system by allocating fewer resources than the sum of resources required by all robots, based on the expectation that robots rarely use all the resources simultaneously.
Oversubscribing can improve resource utilization, but if too many robots contend for resources at once, all robots will experience performance degradation or failures.

\item \textit{Stochastic Algorithmic Latency}: Many robotic algorithms demonstrate stochastic timing behavior, such as asymptotically-optimal rapidly-exploring random tree in motion planning~\cite{ichnowski2019mpt} that relies on random sampling. Large or foundational models are also affected by the stochasticity of the operating system scheduling and buffers~\cite{li2023mathrm, li2023rt}. 
\end{enumerate}

Fault tolerance typically demands an in-depth understanding of the algorithm and specific adaptations against faults, and it involves complex deployment and management processes that require extensive engineering expertise. 
\algname provides fault tolerance for heterogeneous stateless robotic algorithms without requiring ROS2 application modifications. It simultaneously dispatches requests to identical services deployed in multiple cloud data centers and uses the first response received from the replicated services. This model significantly increases the probability of getting timely responses as long as at least one replica and network remains operational and responsive. 
We design a replication-aware routing network that allows resilient and location independent connectivity that persists even if the service is restarted on a different machine.




To reduce the cost of launching
independent 
replicated services, \algname 
can deploy on spot VMs. 
With \algname, even time-sensitive robotics tasks can be executed on those cost-effective servers with per-request fault tolerant characteristics, enjoying the benefits of lower expenses without suffering from the drawbacks of unplanned server shutdown. \algname resiliently manages a pool of spot VMs across multiple cloud and regions, and recreates a new spot VM whenever a replica gets terminated. 

We evaluate \algname on 4 cloud-robotics applications: object detection with YOLOv8~\cite{redmon2016you}, semantic segmentation with Segment Anything (SAM)~\cite{kirillov2023segment}, motion planning with Motion Planning Templates (MPT)~\cite{ichnowski2019mpt}, and a physical pick-and-place task with a UR10e. In experiments,
\algname reduces latency by up to 1.16x in motion planning, including a 5.53x reduction on 99 Percentile (P99), a metric for long-tail latency faults; and \algname reduces average inference latency by 2.1x and P99 latency of object detection by 3.9x under network slowdown. \algname improves SAM's P99 latency by 1.96x when compute resources is contended. 

This paper makes four contributions: (1) an open-source fault-tolerant extension to FogROS2 that enables robots to use replicated, independent cloud resources across different clouds for capacity and availability and stay available as long as one of the replicas and subnetworks is available; (2) cost-effective deployment using spot VMs; (3) cost, fault tolerance, and latency data from an experimental evaluation of \algname on 3 simulated cloud robotics applications; (4) experimental evaluation with a physical robot performing a scan-pick-and-place task.
\section{Related Work}


\paragraph*{Cloud and Fog Robotics}
The use of cloud computing resources for robots, conceptualized as cloud robotics~\cite{kehoe2015survey}, has become increasingly relevant
as large models (e.g., NeRF~\cite{wang2021nerf}, SAM~\cite{kirillov2023segment} and LERF~\cite{Rashid24lifelonglerf} for visual perception) and other computationally demanding algorithms (e.g., MPPI for path planning) are integrated in robotic applications.
Following the Fog Computing paradigm~\cite{bonomi2012fog}, Fog Robotics~\cite{gudi2017fog} utilizes edge resources to improve performance, of cloud computing for a multitude of robotics applications, including grasp planning~\cite{tanwani2019fog}, motion planning~\cite{ichnowski2020fog}, visual servoing~\cite{tian2019fog}, and human-robot interaction~\cite{gudi2018fog}.
In the FogROS2 series of work, we address several concerns of using cloud compute for robotics.
FogROS2~\cite{chen2021fogros} is a cloud robotics framework officially supported by ROS2~\cite{macenski2022robot}.
FogROS2 focused on optimizing for a single cloud and robot using a Virtual Private Network (VPN). Extensions of this work have addressed
the questions of connectivity, latency, and cost.
\sgc (Secure and Global Connectivity)~\cite{chen2023sgc} enables secure communication between distributed ROS2 robot nodes.

\paragraph*{Multi-Cloud Robotics}
FogROS2 is the first multi-cloud robotics that offloads robotics applications to multiple cloud service providers. 
FogROS2-Config~\cite{chen2024fogrosconfig} extends FogROS2 to  navigate available cloud machine selection that meets user-specified time and cost per request.
\algname uses multi-cloud Spot VMs to reduce the cost of the deployment. Spot VMs 
are one such cost-saving purchasing option offered by major cloud service providers that are up to a 90\,\% discount off the standard on-demand pricing, because they can be \emph{preempted}\footnote{\emph{Preempted},  \emph{shut down} and \emph{interrupted} are used interchangeably.} \emph{unpredictably}~\cite{amazon2021spot}. This creates trade-off between the reduced price of spot VMs and having to build an infrastructure handling shutdowns and restarts~\cite{li2016spotpricereview}.
The rates of preemption are highly variable across regions and instance types, from 3\,\% to over 20\,\% chance of preemption per day.
Spot VM prices are also variable, and can fluctuate over the course of a day.
Extensive research has been done to predict spot pricing using statistical models~\cite{javadi2011spotpricemodeling,singh2015spotpricepredict} and learning methods~\cite{baughman2018spotpricelstm,khandelwal2020spotpricerrf}.
Due to their unreliable nature, spot VMs were not considered as part of the machine selection in FogROS2 and FogROS2-Config.



\paragraph*{Fault Tolerance for Latency Sensitive Applications} 
A fault tolerant system is typically based on  \emph{failover}, dynamically switching to one of the machines upon failure of one machine, and \emph{redundancy} by duplicated machines ensuring operation in case of a failure on one system.
FogROS2-LS (Latency Sensitive)~\cite{chen2024fogrosls} implements failover strategy by enabling robots to flexibly connect with one of many servers, but the system takes time to discover and recover from faults by switching to another server that meets latency requirements. 
~\citet{lee2017deepspotcloud} enables fault tolerance on spot VMs for deep learning by actively checkpointing and recovering from the failure. 
On redundancy, \citet{schafhalter2023cloudav} improves the responsiveness of autonomous vehicles by performing operations on both vehicle and cloud. 
On fault tolerance of spot VMs, 
\citet{voorsluys2012reliablespot,poola2016spot,pergentino2019mults} demonstrate spot instances can be used as a cost-efficient setup. However, they focus on the cost optimization on a single server. \citet{alieldin2019spotweb} uses redundancy of spot VMs on web services. However, fault tolerance of continuous and latency-sensitive robotic operations is understudied. With \algname, off-the-shelf robotics applications can run in fault-tolerant environment without the awareness of the robotic application developers.



\section{System Assumptions and Features}

\subsection{System Assumptions}
We assume the application can be partitioned as a \emph{robot} that sends sensor data as requests and awaits control instructions as response from a \emph{service} that encapsulates the algorithm.
We assume that all servers, networks, and faults are independent. 
The services need to be stateless to achieve fault tolerance transparently, or applications needs to use well-defined interfaces from \algname to make the states consistent.

We assume the application is implemented in ROS2, the de-facto platform for building robotics applications. In ROS2, the computational units (\emph{client} and \emph{service}) are abstracted into \emph{ROS2 nodes}. \algname assumes the nodes are connected by ROS2 service communication model~\cite{ros2_service}.

\subsection{Fault Tolerance Properties}

\algname achieves fault tolerance with the following properties: 

\begin{enumerate}
    \item \textbf{Zero Downtime at Faults}. \algname enables fine-grained request-level fault tolerance that ensures a request can be fulfilled as long as at least one replica and network remains operational. 
    \item \textbf{Algorithm-Agnostic}. \algname operates independently of the specific algorithms used in applications as long as it is stateless.
    \item \textbf{Failure Cause-Agnostic}. \algname does not need to be tailored to specific failure, and remains functional as long as at least one service is available and connected.
    \item \textbf{Hardware-Agnostic}. \algname is agnostic to heterogeneous hardware resource options provided by the cloud, supporting simultaneous use of different resource types and fault tolerance level. 
    \item \textbf{Multi-cloud, Multi-region}. Since the failure may occur to a specific region or cloud provider, \algname, as part of the Sky Computing paradigm~\cite{chasins2022sky}~\cite{yang2023skypilot}, offers a unified interface for interacting with various cloud service providers and uses cloud computing resources across different clouds simultaneously. 
\end{enumerate}

\subsection{\algname Failure Qualification}
Multiple cloud servers can provide fault tolerance to region or server-specific compute or network failures. 
We assume the robot has a persistent and stable connection to at least one of the cloud servers. 

The probability of a VM failing at any moment in time is
\[
P_{\text{VM}_i}(\text{failure}) =
\frac{\text{recovery\_time}_{\text{VM}_i}}{\text{uptime}_{\text{VM}_i} + \text{recovery\_time}_{\text{VM}_i}}, 
\]
and the probability of a system failure with $N$ spot VMs is
\[
P_{\text{system}}(\text{failure}) = \prod_{i \in N} P_{\text{VM}_i}(\text{failure}).
\]
Given a desired maximum failure rate for the system and the failure rate for all VMs used, we can calculate the required number of VM replicas as
   \[  N = \left\lceil \frac{\log{P_{\text{VM}}(\text{failure})}}{\log{P_{\text{system}}(\text{failure})}} \right\rceil. \]

To reduce the probability of failure, one can either increase the number of replicas, or use a combination of spot and on-demand instances. 
For example, a spot VM is preempted every 15 hours on average from experiments by Skypilot paper~\cite{yang2023skypilot} and re-creating and initiating a new instance with FogROS2 can take up to 20-minute downtime~\cite{chen2021fogros}, the probability of simultaneous preemptions with two spot VMs is less than 0.05\,\%, fulfilling the Service Level Agreement of AWS on non-spot VMs~\cite{aws_sla}. 


\subsection{New Features}

\algname is distinguished from previous work with the following features: 
\begin{enumerate}
    \item \textbf{Cost-effectiveness}.  \algname can provide significantly lower cost by using spot VMs. 
    \item \textbf{Scalability} \algname allows robots to launch a adjustable 
    number of machines to reduce the probability that all the replica service deployments are not available. 
    \item \textbf{Flexibility} \algname is flexible to the choice of available hardware resources. For example, one can mix low-specification on-demand cloud machines for up-time with high-specification spot VMs for speed. 
    \item \textbf{Adaptive and Resilient recovery} \algname automatically recovers from service interruptions, preserving the intended level of fault tolerance automatically.
\end{enumerate}

\section{\algname Design}
This section describes how \algname (1) achieves transparent fault tolerance for ROS2 applications and (2) resiliently maintains a pool of cost-effective cloud servers.

\subsection{Overview}
\begin{figure}
    \centering
    \includegraphics[width=0.9\linewidth]{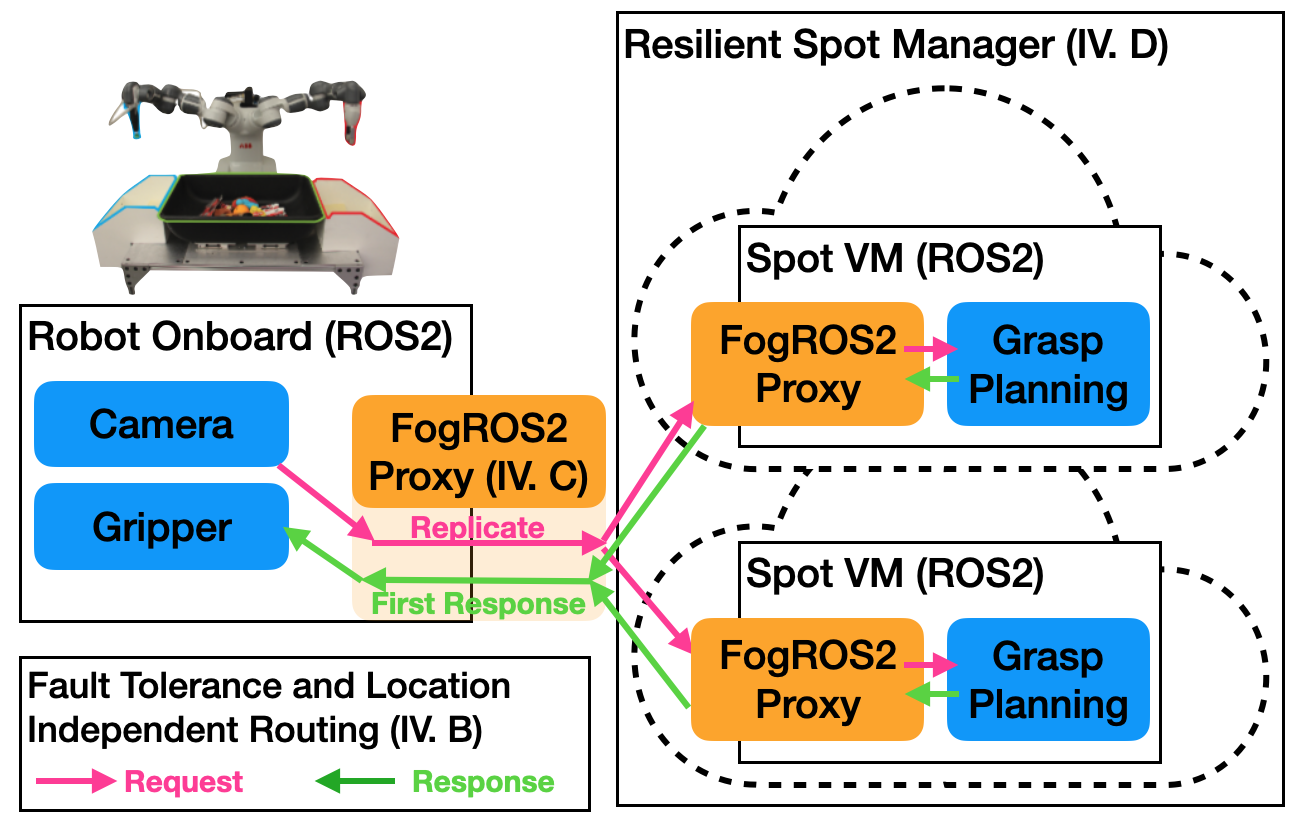}
    \caption{\textbf{System Overview of \algname}  \algname transparently proxies ROS2 communication. It sends requests to multiple replicated spot VMs, and routes the first response back to the robot. \algname manages spot VMs to resiliently recover from unpredictable terminations. }
    \label{fig:design:overview}
\end{figure}

Figure \ref{fig:design:overview} shows an overview of how \algname achieves both fault tolerance and cost effectiveness. \algname sends requests to multiple replicated spot VMs, and routes the first response back to the robot. It resiliently manages spot VMs to  recover from unpredictable terminations.   

\textbf{Interface}.
A user interfaces to \algname is through extensions to the ROS2 launch system.
The interface is identical to standard ROS2 launch scripts, other than specifying the hardware requirements and desired fault tolerance level.
We embrace the Sky \cite{chasins2022sky} multi-cloud paradigm
that user can directly specify generic hardware requirements instead of specific cloud machine types, which are compatible across heterogeneous cloud service providers. 


\subsection{Application-Agnostic Fault Tolerance through Replication}


To launch (unmodified) ROS2 nodes in the cloud,
\algname 
is a \textit{multi-cloud launcher} that facilitates cloud initialization  and \textit{a replication-aware proxy} that connects ROS2 robot client and cloud service with a global fault-tolerant network. 

The multi-cloud launcher initializes cloud servers with a ROS2 environment, and provisions secure communication through \algname \emph{proxies}. 
To local ROS2 network, the proxy serves the requests as local ROS2 service and interacts with nodes (e.g., sensors and controllers) on the robot as if the ROS2 service nodes on the cloud were all on the same local computer.
On a new request, the proxy of \algname sends the request to all replicated service nodes. When the proxy receives responses from the replicas, it passes only the first response to the robot. This is agnostic to applications and causes of the fault. As long as one response comes back to the robot, the request is fulfilled. To the cloud server, the proxy receives the request from the network, and invokes the ROS2 service on the cloud. 

\textbf{Multi-Cloud Fault Tolerant Launching Process}.
Initializing fault tolerant cloud robotics includes the following steps (1) The robot provisions multiple Cloud servers across different regions and data centers. \algname can automatically select the region based on network latency and cloud operating cost. The user can override the selection with configuration. \algname uses SkyPilot~\cite{yang2023skypilot} to interface with heterogeneous cloud service providers. (2) The robot initializes all the cloud servers with ROS2 and the robot service application dependencies. FogROS2~\cite{ichnowski2023fogros} details the initialization procedure. (3) Robot and all Cloud servers generate and share communication security credentials (4) Robot and all Cloud servers run \algname proxy 
(5) All proxies discover each other automatically, establish global and resilient connectivity with the generated security credentials and desired topology. 

By default, all robotics services 
have at least two replicas at different cloud data centers.
Critical services can use more replicas and adapt based on changing conditions (such as time of the day, budget). Without terminating the existing cluster, one can use Command Line Interface (CLI) to scale up and down the number of replicas dynamically:
\[
\texttt{\footnotesize ros2 fog scale [up/down] [args]}
\]



\begin{figure}
    \centering
    \includegraphics[width=\linewidth]{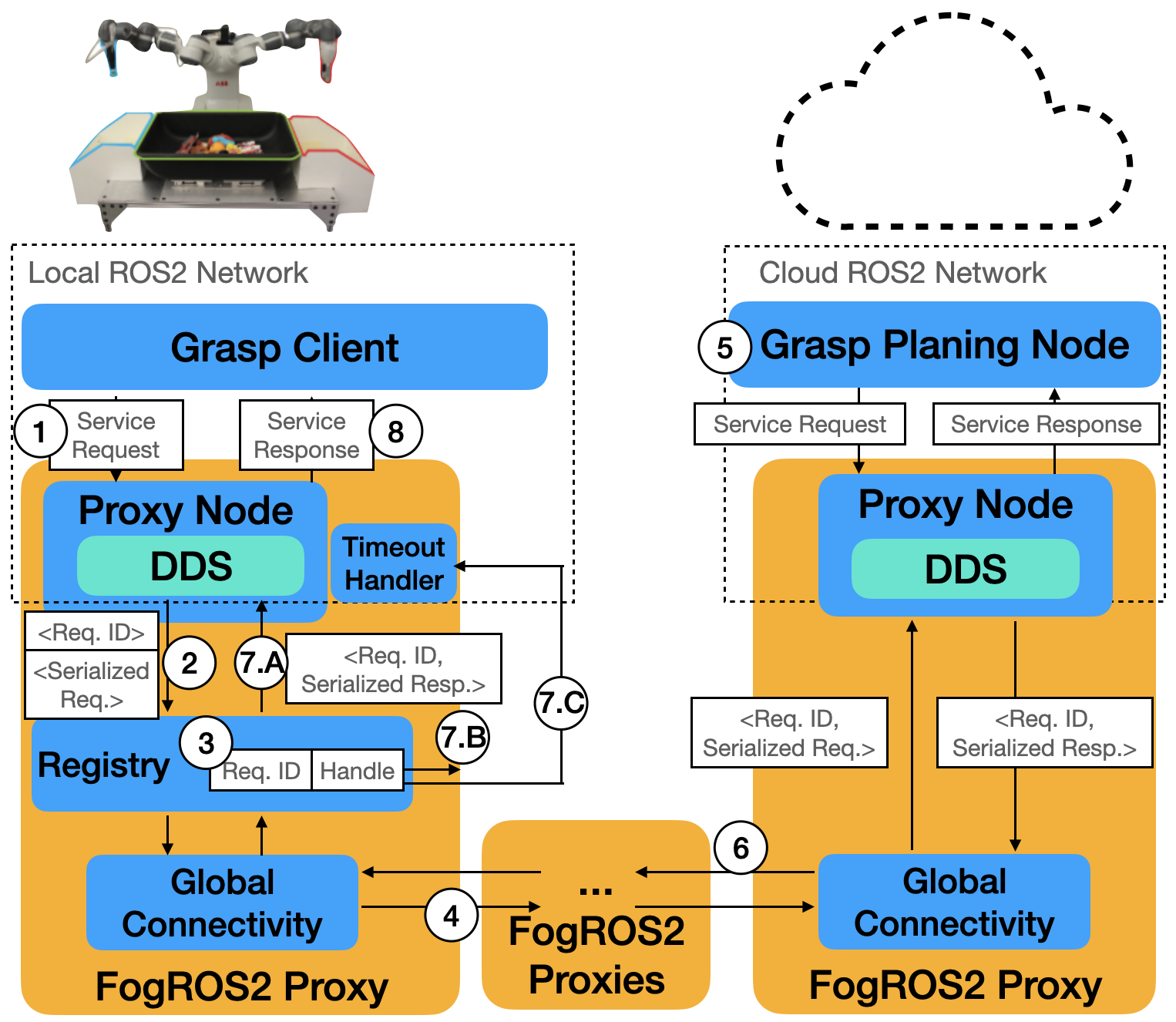}
    \caption{\textbf{Flow diagram of \algname on handling new requests} 
    The \algname replication-aware proxy handles ROS2 grasp planning request with fault tolerance guarantees with multiple steps. (1) The ROS2 application sends a request on the local ROS2 network.
(2) The proxy running on the robot receives the request and extracts the content and a unique identifier 
from ROS2 middleware (rmw) layer buffer. 
(3) The proxy registers the unique ID from rmw with the \textit{handle}, which includes a callback function if the response of the request arrives and a callback function for timeout.
(4) The proxy securely sends the request to proxies running on replicated Cloud machines. 
There can be multiple proxy hops between the robot and the server that hosts the desired ROS2 service. 
The request message carries the unique identifier and the proxy adds an entry in the registry table.
(5) The proxy running on the cloud converts the message to a standard ROS2 request message, and invokes the ROS2 service on the cloud and gets the response. 
(6) The proxy on the cloud sends the response back to the proxy on the robot. 
(7.A) The robot checks if it handled the response 
with the unique identifier; 
(7.B On the duplicated responses) The proxy drops the response if it was already handled. 
(7.C On timeout) The proxy calls the timeout handler (such as returns with empty response) and cleans up the registry table.
(8) The robot sends the response to the application on the robot through standard ROS2 protocol. 
}
    \label{fig:design:flow}
    \vspace{-10pt}
\end{figure}

\textbf{Replication-Aware Fault Tolerant Connectivity}.
Figure \ref{fig:design:flow} shows the workflow of \algname on handling requests with fault tolerance.  A ROS2 application sends a request via the local ROS2 network, where the robot's proxy captures, extracts its serialized content a unique identifier from the ROS2 middleware layer (rmw). The proxy stores the identifier with a \textit{handle} and sends the request to the cloud. The handle incorporates a callback function called on response arrival and on timeout. The cloud proxy invokes the corresponding cloud-based ROS2 service to compute and send the response back. The robot's proxy verifies the response with the unique identifier, then either marks the request as completed and sends back to the robot, or discards as a duplicate.


\subsection{Resilient and Flexible Connectivity}
The fault tolerance workflow of \algname is established on a global peer-to-peer network, a fabric formed by interconnected proxies. The connections are resilient to service changes and flexible to various network topologies. 

When a cloud machine is interrupted and rebooted, \algname launches another cloud virtual machine. The launched cloud machine is typically different cloud machine with changed network connectivity information (such as IP address). 
The connections between proxies should be resilient to such interruption that even if the server gets interrupted and relaunched at a different physical machine, it can still maintain its connection to the robot. 

\algname achieves this by assigning a globally unique identifier to every peer-to-peer connection between proxies. 
This identifier can be generated deterministically between the proxies that no other ROS2 service can produce the same identifier. \sgc~\cite{chen2023sgc} provides details about the identifier construction process and guarantees.  
In this work, we extend the identifier of \sgc to network connections, such that even if the original machine is interrupted and restarted at a different place, it can still deterministically generate the identifier and resume the connectivity. 
\algname connects the proxies on the robot and on the cloud with the same globally unique identifier by a metadata server. The metadata server exchanges network information, monitors the status of the connections, and cleans up the connectivity information if one of the proxy reports its peer as disconnected. The metadata server can be hosted on lightweight, low-bandwidth and accessible cloud servers. The server only facilitates connectivity, and no application data flows through the server, so a failure of the server does not lead to a system failure. 

\begin{figure}
    \centering
    \includegraphics[width=0.9\linewidth]{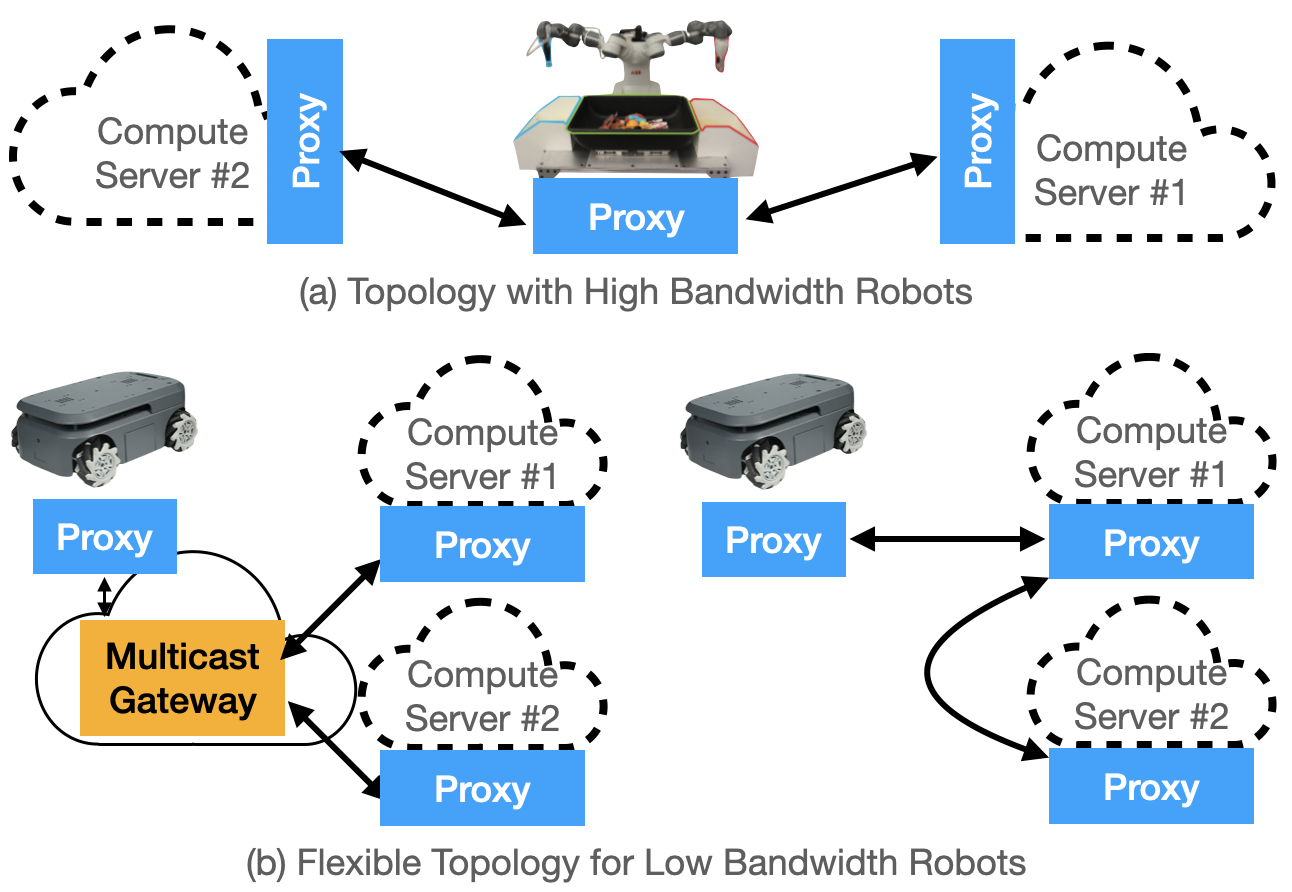}
    \caption{\textbf{Flexible Topology for Different Bandwidth of Robots} \textbf{(a)} Since \algname sends replicated requests to multiple cloud machines, it demands more network bandwidth than conventional cloud-robotics deployments. \textbf{(b)} \algname allows flexible topology so that low-bandwidth robot can leverage cloud machines with higher bandwidth to forward to replicated services. One can either use dedicated gateway machine (left) or existing compute servers (right). }
    \label{fig:mcast}
    \vspace{-10pt}
\end{figure}

\textbf{Flexible and Scalable Topology For Bandwidth-Limited Robots}.
We consider mobile robots with low network bandwidth that may be bottlenecked if sending the request to multiple cloud servers. In this case, we need an intermediate server with higher bandwidth as gateway to forward the requests to other proxies. The server can be on the cloud or edge.
The transport of \algname is not constrained to having the client and server be directly connected. In \algname, the proxy can be in a tree-like structure, where each edge the tree connects between an intermediate hop and the proxy. The proxy finds and establishes connections with its peers by flattening the tree. Figure \ref{fig:mcast}\,(b) shows two possible topology examples, which one can launch a lightweight gateway server to facilitate the message replication, or directly use a proxy on the compute service node. 

\subsection{Resilient and Cost-Efficient Spot VMs}
\algname 
can use spot VMs to reduce the cost of the replicated cloud server deployment. 
Since spot VMs can be up 90\,\% lower cost than their equivalent on-demand VMs,
running two spot VMs at the same time can still enjoy as much as 80\,\% cost reduction. 
Earlier studies~\cite{10.1145/3458336.3465301} show a bioinformatics task with 24 spot VMs on Google Cloud Platform experiences a preemption every 36 minutes on average. 
Because of the fault tolerance extension of FogROS2, we are able to guarantee per-request availability that as long as there is one spot VM not preempted and available to serve the request. 
\algname launches and manages spot VMs across multiple cloud service providers and regions. 
\algname regularly monitors the status of the spot VMs. When a spot VM is interrupted, \algname re-launch the service node on a new spot VM and re-establishes connectivity.

\section{Experiments}

\subsection{Setup}
Without modifying the application code, we apply \algname to three cloud robotics applications: visual object detection with YOLOv8~\cite{redmon2016you}, Semantic Segmentation with Segment Anything~\cite{kirillov2023segment} and motion planning with Motion Planning Templates (MPT)~\cite{ichnowski2019mpt}. 
We use SkyPilot \cite{yang2023skypilot} to select servers given a hardware specification. 
For all the experiments, we use Amazon Web Services (AWS) spot VMs with two replicas in different regions, us-west-1 (California) and us-west-2 (Oregon). The workstation connects with spot servers with Ethernet connection.

\textbf{Metrics}.
We compare \algname with baseline single-server deployment by average latency collected by 100 trials for motion planning and 300 trials for robot vision tasks. We quantify \textit{long-tail} anomalous latency faults with 99 Percentile (P99) latency, the runs with the top 1\,\% latency. 

\textbf{Cloud Cost} 
Table \ref{tab:eval:cost} shows the US\$ per hour cost of \algname compared to typical cloud robotics single-server setup. With Spot VMs, \algname is up to 2.13x cheaper than conventional single-server cloud robotics setup.

\begin{table}[t]
    \centering
    \footnotesize
    \scalebox{0.9}{
    \begin{tabular}{lcccccc}
    \toprule
            & Server & Single Server & \multicolumn{3}{c}{\algname (USD per Hour)}\\
    Application & Core &  (USD per Hour) & On-Demand & & {Spot VM} \\
    \midrule
    YOLO & 16 & 0.40 (1.17x) & 0.80 (2.34x) & & \textbf{0.34} (1x) \\
    SAM & 4$^*$ & 0.53 (1.21x) & 1.06 (2.42x) & & \textbf{0.44} (1x) \\
    MPT-Home & 32 & 1.79 (1.05x) & 3.58 (2.10x) & & \textbf{1.69} (1x) \\
    MPT-Cubicles & 32 & 1.79 (1.05x) & 3.58 (2.10x) & & \textbf{1.69} (1x) \\
    MPT-TwistyCool & 64 & 3.58 (2.13x) & 7.16 (4.26x) & & \textbf{1.68} (1x) \\
    Pick-Scan-Place & 16 & 0.40 (1.17x) & 0.80 (2.34x) & & \textbf{0.34} (1x) \\
    \bottomrule
    \end{tabular}
    }
    \caption{\textbf{\algname hourly cost (USD per Hour) comparison to a Single-Server (one on-demand cloud machine)} \algname can use two on-demand machines deployment or spot VMs. In SAM, we use Nvidia T4 GPU accelerator for the inference. \algname is cheaper to run 2 spot VMs as opposed to a single server deployment up to 2.13x. }
    \label{tab:eval:cost}
    \vspace{-10pt}
\end{table}

\subsection{Case Study: Parallel Motion Planning}
We perform a latency analysis of \algname for parallel motion planning tasks on 3 different scenarios of varying complexity provided by Open Motion Planning Library (OMPL)~\cite{OMPL}. Given a robot's initial state and goal state, the motion planner computes the waypoints required to move the robot to the goal while avoiding the obstacles. The motion planning algorithms we test are iterative optimization problems with stochastic solve times.

\begin{figure}
    \centering
    
    \includegraphics[width=\linewidth]{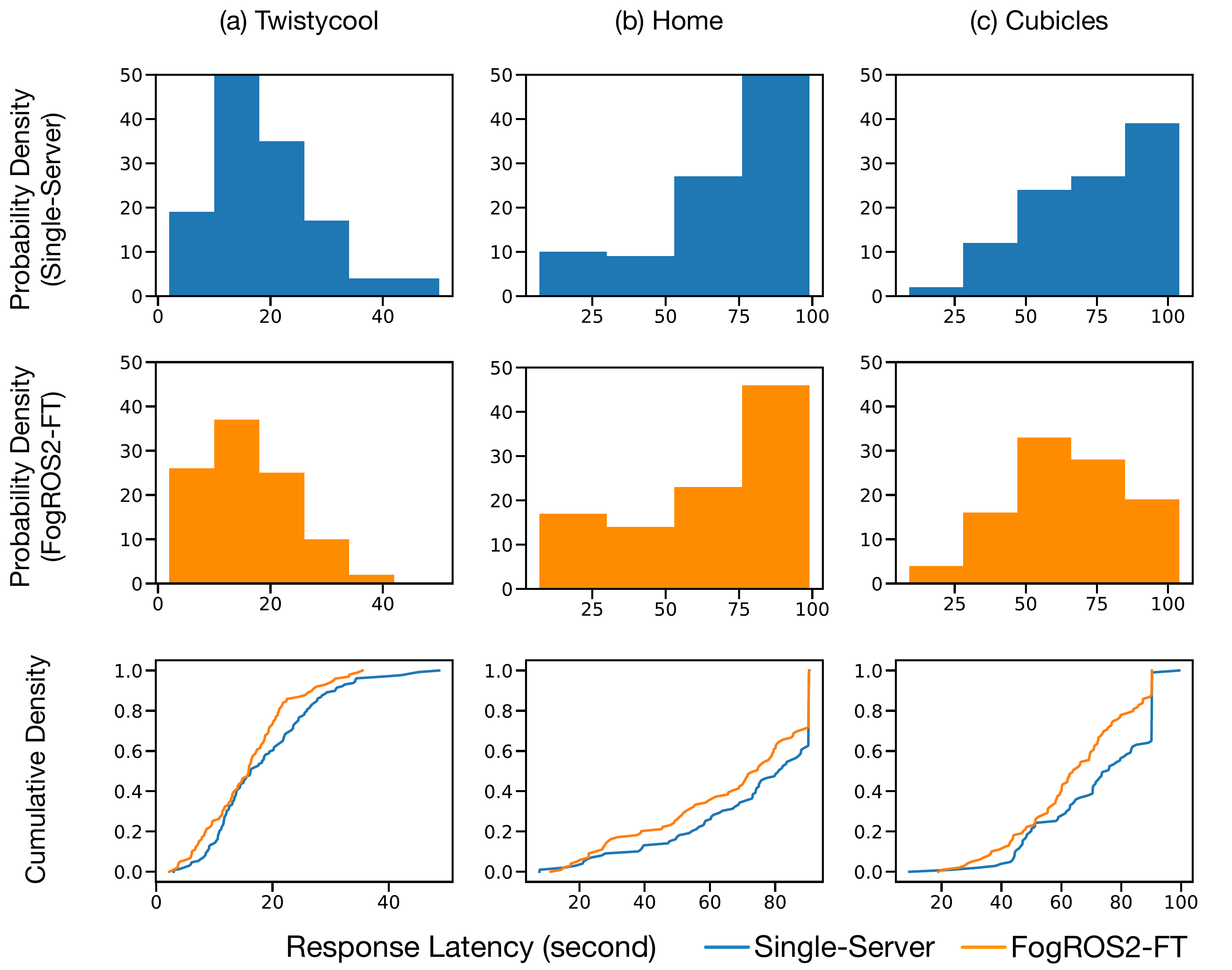}
    \caption{\textbf{\algname Latency on Motion Planning Template} We tested \algname on 3 different motion planning environments (columns (a), (b), and (c)). Due to the stochastic nature of the algorithm, we aggregated results for each scenario and server configuration over 100 trials with a 100\,s timeout. The \textbf{(top row)} 
    shows the frequency histogram for the scenario when {run} with Single-Server (in blue).
    The \textbf{(middle row)} shows the frequency histogram for the scenarios when run on 2 servers with \algname (in orange). With all scenarios, the shift left of \algname histograms (in orange) relative to their corresponding single-server histograms (in blue) indicates improved latency performance when running on replicated servers. The \textbf{(bottom row)} compares the cumulative distribution functions (CDF) for single-server (in blue) and two servers (in orange). The two-server CDF is left relative to the single-server CDF indicating an overall improved performance with lower average latency for all scenarios. } 
    \label{fig:eval:mpt}
    \vspace{-10pt}
\end{figure}

\textbf{Latency Analysis}
We compare the latency probability between single-server and \algname with 2 servers in Fig. \ref{fig:eval:mpt}.
The results for the motion planning latency analysis are summarized in Figure~\ref{fig:eval:mpt}. For all 3 scenarios, \algname reduces average latency by up to 1.22x on Cubicles. \algname significantly mitigates anomalous long and randomized latency by 1.42x (TwistyCool) and 5.53x (Cubicles) on P99 long-tail latency, because the probability of simultaneous anomalous high latency is rare. 


    

\subsection{Case Study: Robot Vision with YOLOv8 and SAM}


\begin{figure}
    \centering
    \includegraphics[width=\linewidth]{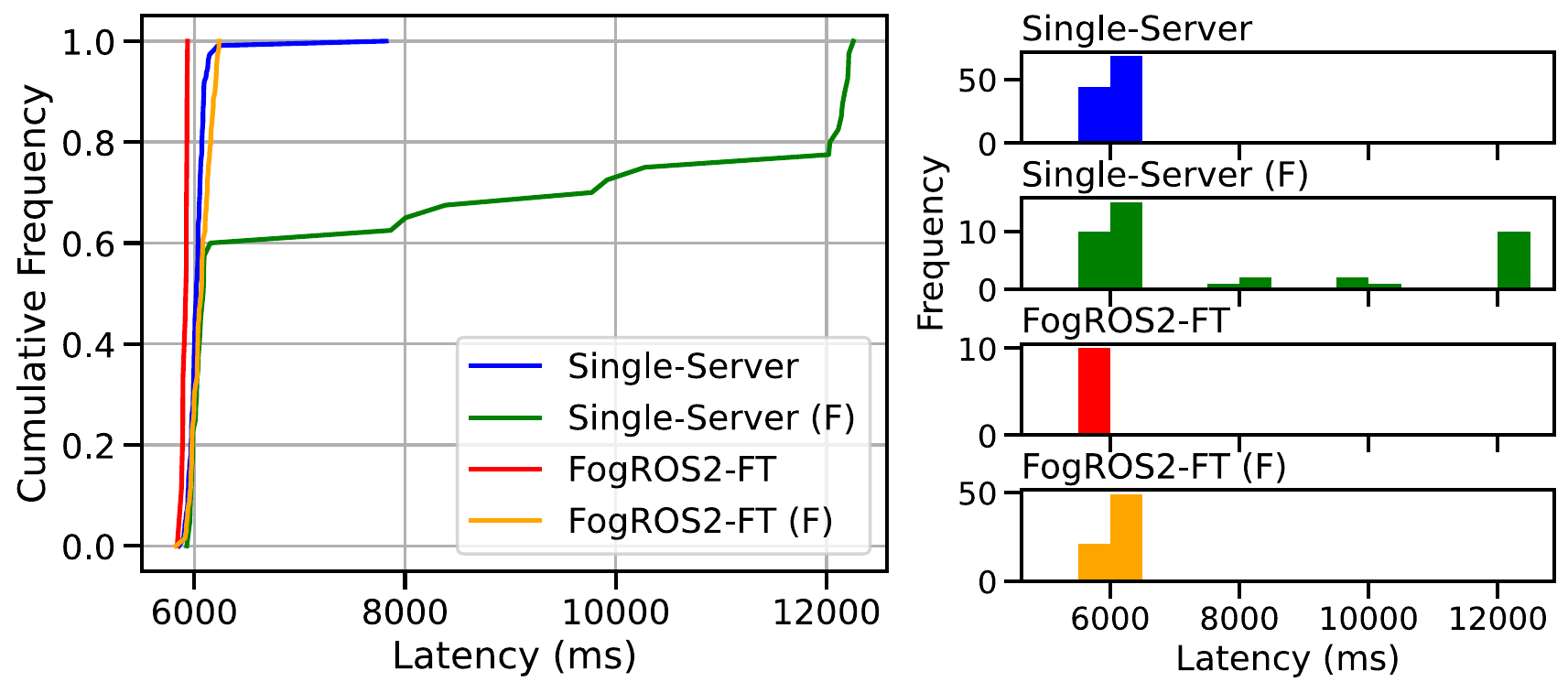}
    \caption{\textbf{\algname on Semantic Segmentation with SAM under Faults (F) by Compute Resource Oversubscription}  \algname improves 30\,\% of the long-tail P99 latency for SAM. 
    As cloud GPU resources are typically oversubscribed and shared by multiple concurrent clients, we emulate such resource contention by running another periodic and concurrent client that alternates to send requests to \algname server. \algname significantly reduces the long-tail latency caused by the resource contention by 1.96x. }
    \label{fig:eval:sam}
    \vspace{-10pt}
\end{figure}

\begin{figure}
    \centering
    \includegraphics[width=\linewidth]{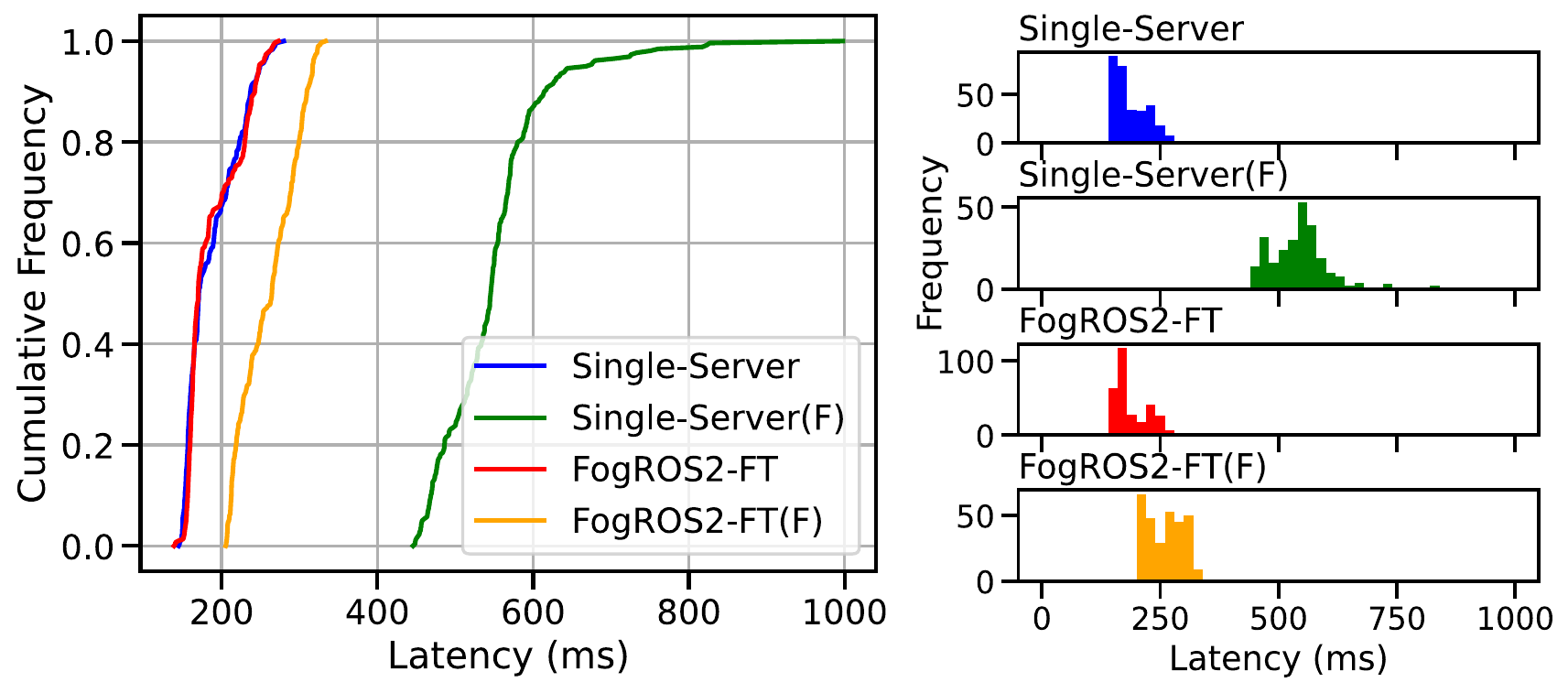}
    \caption{\textbf{\algname on Object Detection with YOLOv8 under Faults (F) by Regional Network Slowdown} 
    We use \algname to launch a multi-cloud cluster that includes us-west-1 (California) and us-west-2 (Oregon).
    \algname and single-server deployments demonstrates similar latency under good network conditions, as the additional latency of sending to another replicated service is amortized \algname using the first response. 
    To simulate network congestion that can occur when a robot has trouble connecting to a specific data center,
    we introduce 100\,ms latency on all our servers in us-west-1 (labeled with (F) for `fault'). With the network slowdown, \algname is 2.1x faster on average than non-fault tolerant deployment, and 3.9x faster on P99 latency.}
    \label{eval:yolo:results}
    \vspace{-10pt}
\end{figure}


We evaluate \algname with SAM and YOLOv8. SAM is a computationally expensive algorithm that requires long computational time even with GPU. The result in Fig. \ref{fig:eval:sam} (Single-Server) shows the latency of running SAM on a typical setup with one cloud server, SAM shows a long-tail and significant P99 latency. \algname improves the P99 latency by 1.3x by selecting the first response from the duplicated services. 
Compared to SAM, YOLO has lower inference latency, thus more network-intensive. 
In this case, \algname demonstrates similar latency as Single-Server deployment for the tradeoff that \algname consumes more network bandwidth but the additional latency is amortized by more stable response time.

\textbf{Faults at Compute Resource Oversubscription} 
We evaluate \algname against the faults of resource contention if many robots are contending on few resources, where one may choose to run multiple services on the same physical machine. By comparing faults (F) between Single-Server (F) and \algname (F) in Fig.\,\ref{fig:eval:sam}, \algname reduces the latency by 1.31x and long-tail P99 latency by 1.96x. 

\textbf{Faults at Regional Network Slowdown}. 
Sometimes robot may experience slowdown when connecting to a specific cloud data center. This can be caused by a periodic and regional network congestion or physical location. By comparing faults (F) between Single-Server (F) and \algname (F) in Fig.\,\ref{eval:yolo:results}, \algname reduces the latency by 2.1x and long-tail P99 latency by 3.9x. 

\subsection{Case Study: Physical Scan-Pick-and-Place}


\begin{figure}
    \centering
        \centering
        \includegraphics[width=\linewidth]{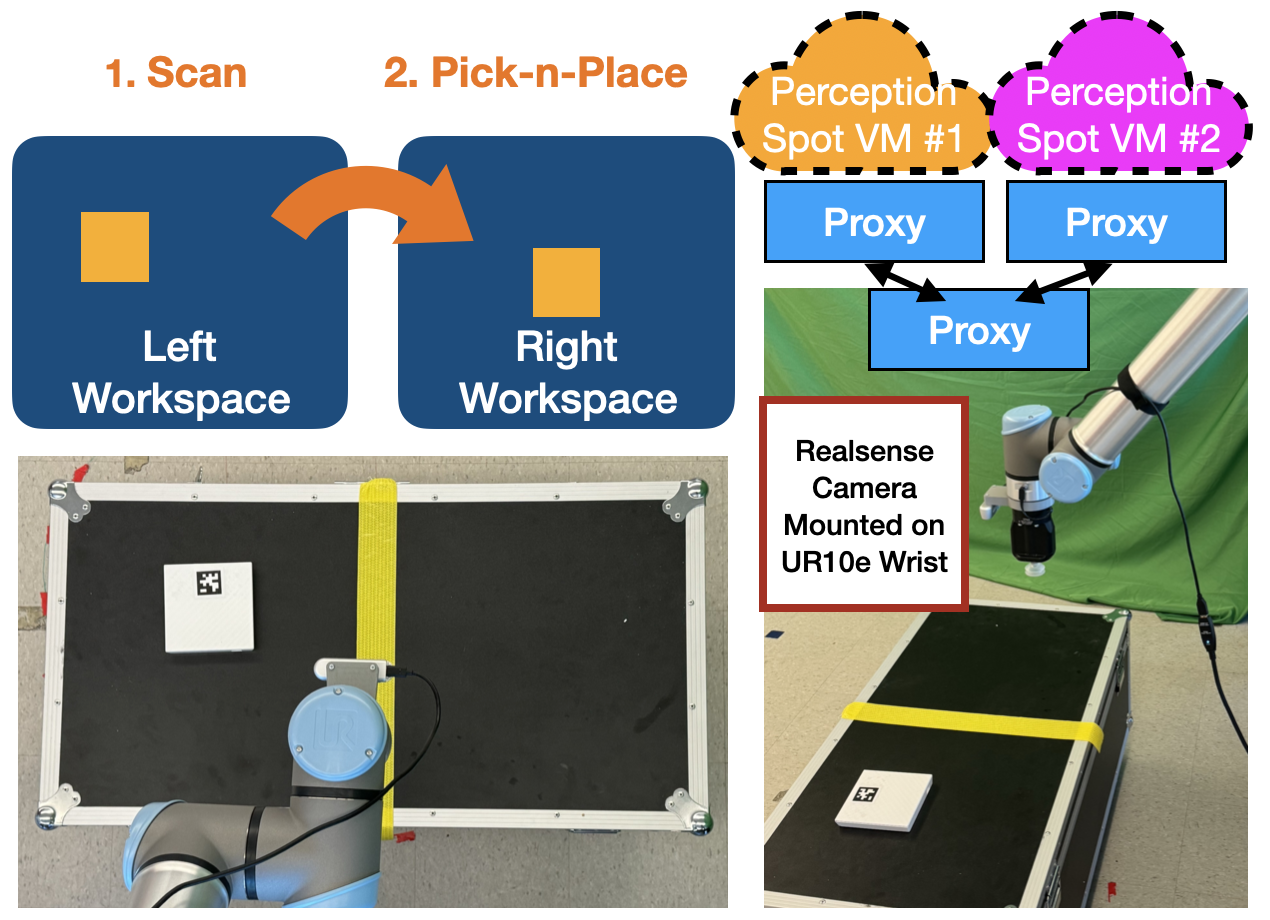}
        \caption{\textbf{Physical Setup and System Diagram.} \textbf{(Left)} In the scan-pick-and-place robotic task, the goal is to move an object from one workspace to a randomly generated location on the other workspace.  We repeat the task (from left to right, from right to left) for hours to show the continuity and robustness of \algname in a physical robot system. \textbf{(Right)} The system diagram has two spot VMs providing Apriltag localization services to a UR10e robot with a RealSense Camera mounted on its wrist. The robot and camera connect to an edge computer that runs a robot motion planner and a \algname proxy to connect to the replicated cloud services.}
        \label{fig:eval:pnp:setup}
        \vspace{-10pt}
\end{figure}

\begin{figure}
    \includegraphics[width=\linewidth]{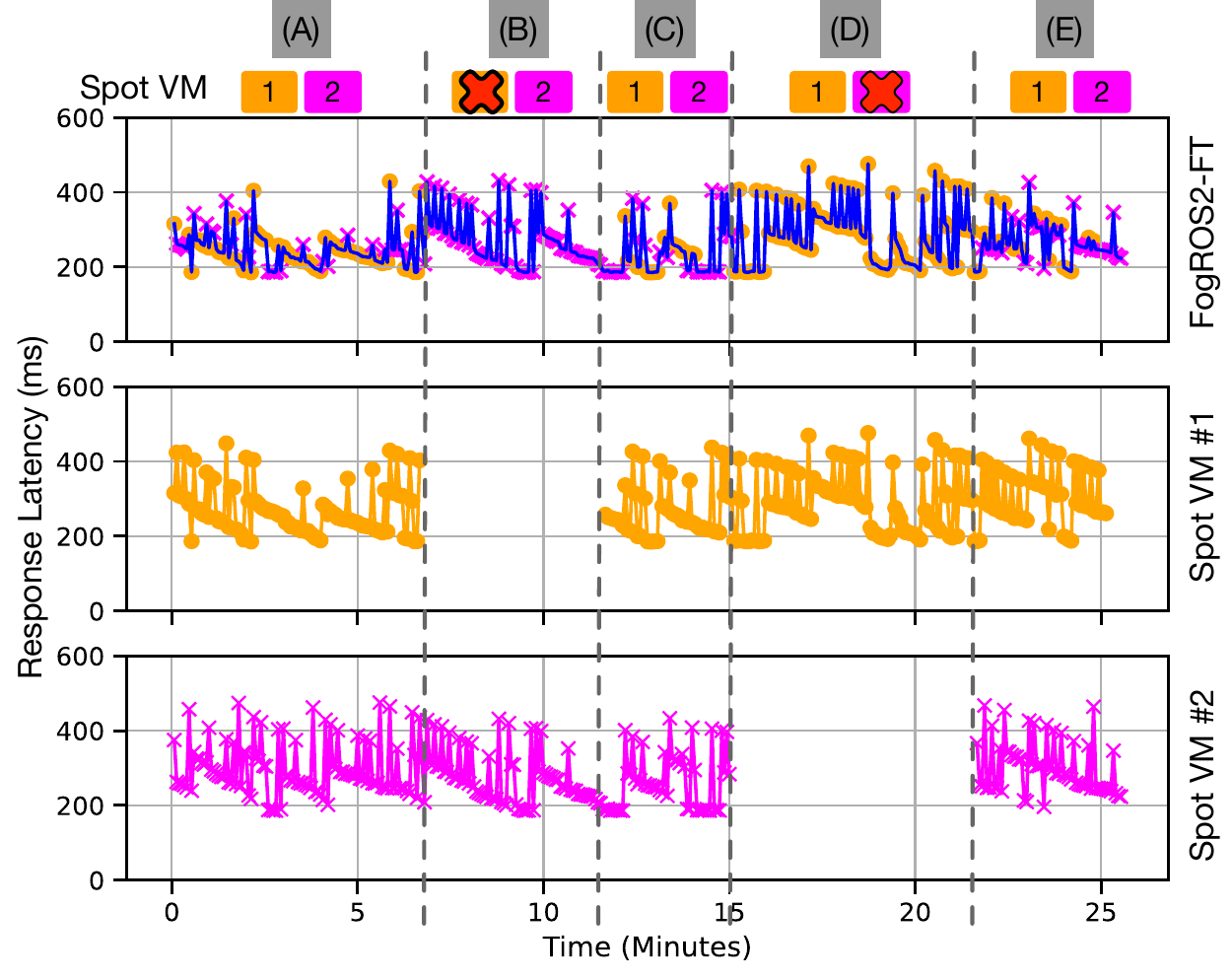}
    \caption{\textbf{\algname on Physical Fault Tolerant Pick-Scan-Place} (A) \algname achieves reliable latency by choosing the first response with two active servers (B) We manually terminate one spot VM to emulate a fault, and \algname uses the responses from spot VM \#2. (C) \algname automatically recovers spot VM \#1 from failure by re-initializing user's environment and connectivity (D) We terminated spot VM \#2 to emulate spot VM preemption (E) \algname recovers and uses responses from both servers. 
    The recovery downtime can be improved by tools from FogROS2~\cite{ichnowski2023fogros}.}
    \label{fig:eval:physical:result}
    \vspace{-10pt}
\end{figure}

We evaluate \algname with cloud-based scan-pick-and-place with a fundamental physical robotic skill for many robot tasks, such as random bin picking, sorting, kitting, and conveyor belt pick-n-place. 
By offloading visual perception services, \algname improves the responsiveness and reliability in spite of faults.


\textbf{Experiment Setup}
Fig. \ref{fig:eval:pnp:setup} shows the physical setup of the Scan-Pick-and-Place evaluation. 
The Universal Robots arm (UR10e) is mounted with an Intel RealSense D435i on the wrist. 
We use a suction gripper to grasp a plastic CD player marked with an Apriltag. 
The system executes a cyclical procedure that begins with the robot moving to a predefined joint position. 
We implement a perception ROS2 service that 
 uses Apriltag~\cite{wang2016apriltag} for pose estimation.
The ROS2 service takes in 640x480 resolution image frames and returns a 6D pose of the target, and the robot picking motion is generated and executed with the MoveIt2 motion planning tool on the edge computer, completing the feedback loop. 
The robot arm alternate to pick-and-place a box between the left section and the right section of a designated platform.


\textbf{Latency Analysis} 
We conducted the scan-pick-and-place repeatedly and manually interrupt the spot VMs to demonstrate the robustness of the system.
Fig.\,\ref{fig:eval:physical:result} shows the latency timeline of \algname on Physical Fault Tolerant Pick-Scan-Place. We manually terminated one of the two spot instances (scenario B and D). 
There are two takeaways: (1) \algname can achieve reliable latency by choosing the first response with two active servers  (2) \algname can recover from server failures, such as termination. 
As long as one of the services is available and responsive, it can provide continuous operation for latency-sensitive applications.

\section{Conclusion}

In this work, we explore concurrent execution across identical and stateless service deployments on different cloud machines with \algname, and use the first received response to achieve the tolerance against independent faults. Evaluation shows \algname can reduce the long-tail latency by up to 5.53 times. 
It can be deployed on cost-efficient spot VMs that the fault tolerant system can be up to 2.1x cheaper than conventional cloud robotics setup.

In future work, we will use multi-path connection, for example, having mobile robot to use 5G cellular connection, Wi-Fi and Ethernet connection simultaneously to avoid failures on single network connection. This resolves the assumption of \algname that servers and faults to be independent and prevents the same network link between robot and cloud to be the single point of failure.

\section*{Acknowledgements}
This research was performed at the AUTOLAB at UC Berkeley in affiliation with the Berkeley AI Research (BAIR) Lab. The authors were supported in part by donations from Robert Bosch Research. Cloud credits for experiments are provided by Amazon AWS.





\renewcommand*{\bibfont}{\footnotesize}
\printbibliography
\end{document}